\documentclass{article} 
\usepackage{nips13submit_e,times}
\usepackage{url}

\usepackage{epsfig}
\usepackage{graphicx}
\usepackage{amsmath}
\usepackage{amssymb}
\usepackage{algorithm,algorithmicx,algpseudocode,url,verbatim}

\usepackage{array}

\usepackage[pagebackref=true,breaklinks=true,colorlinks,bookmarks=false,urlcolor=blue]{hyperref}

\newcolumntype{M}{ >{\centering\arraybackslash} m }

\title{Learning Factored Representations in a \\ Deep Mixture of Experts}

\author{
David Eigen $^{1,2}$
\quad
Marc'Aurelio Ranzato $^{1}$
\thanks{Marc'Aurelio Ranzato currently works at the Facebook AI Group.}
\quad
Ilya Sutskever $^1$
\\
$^1$ Google, Inc. \\
$^2$ Dept. of Computer Science, Courant Institute, NYU \\
{\tt deigen@cs.nyu.edu} \quad {\tt ranzato@fb.com} \quad {\tt ilyasu@google.com}
}

\newcommand{\eqn}[1]{Eqn.~\ref{eqn:#1}}
\newcommand{\fig}[1]{Fig.~\ref{fig:#1}}
\newcommand{\tab}[1]{Table~\ref{tab:#1}}
\newcommand{\secc}[1]{Section~\ref{sec:#1}}

\def\etal{{\textit{et~al.~}}}
\def\ie{{\textit{i.e.~}}}
\def\eg{{\textit{e.g.~}}}

\newcommand\softmax{{\rm softmax}}

\nipsfinalcopy 

\begin{document}

\maketitle

\begin{abstract}

Mixtures of Experts combine the outputs of several ``expert'' networks, each of
which specializes in a different part of the input space.  This is achieved by
training a ``gating'' network that maps each input to a distribution over the
experts.  Such models show promise for building larger networks that are still
cheap to compute at test time, and more parallelizable at training time.  In
this this work, we extend the Mixture of Experts to a stacked model, the
\emph{Deep Mixture of Experts}, with multiple sets of gating and experts.  This
exponentially increases the number of effective experts by associating each
input with a \emph{combination} of experts at each layer, yet maintains a
modest model size.  On a randomly translated version of the MNIST dataset, we
find that the Deep Mixture of Experts automatically learns to develop
location-dependent (``where'') experts at the first layer, and class-specific
(``what'') experts at the second layer.  In addition, we see that the different
combinations are in use when the model is applied to a dataset of
speech monophones.  These demonstrate effective use of all expert
combinations.

\end{abstract}

\section{Introduction}

Deep networks have achieved very good performance in a variety of tasks, \eg
\cite{Kriz12, Graves13, Ciresan11}.  However, a fundamental limitation of these
architectures is that the entire network must be executed for all inputs.  This
computational burden imposes limits network size.  One way to scale these
networks up while keeping the computational cost low is to increase the overall
number of parameters and hidden units, but use only a small portion of the
network for each given input.  Then, learn a computationally cheap mapping
function from input to the appropriate portions of the network.

The Mixture of Experts model \cite{Jacobs91} is a continuous version of this: A
learned gating network mixes the outputs of $N$ ``expert'' networks to produce
a final output.  While this model does not itself achieve the computational
benefits outlined above, it shows promise as a stepping stone towards networks
that can realize this goal.

In this work, we extend the Mixture of Experts to use a
different gating network at each layer in a multilayer network, forming a Deep
Mixture of Experts (DMoE).  This increases the number of effective experts by
introducing an exponential number of \emph{paths} through different
\emph{combinations} of experts at each layer.  By associating each input with
one such combination, our model uses different subsets of its units for
different inputs.  Thus it can be both large and efficient at the same time.

We demonstrate the effectiveness of this approach by evaluating it on two datasets.
Using a jittered MNIST dataset, we show that the DMoE learns to factor
different aspects of the data representation at each layer (specifically,
location and class), making effective use of all paths.  We also find that all
combinations are used when applying our model to a dataset of speech monophones.

\section{Related Work}

A standard Mixture of Experts (MoE) \cite{Jacobs91} learns
a set of \emph{expert} networks
$f_i$ along with a \emph{gating} network $g$.  Each $f_i$ maps the input $x$ to
$C$ outputs (one for each class $c = 1, \dots, C$), while $g(x)$ is a
distribution over experts $i = 1, \dots, N$ that sums to 1.  The final output is then given by \eqn{moe-f}
\begin{eqnarray}
\label{eqn:moe-f}
F_{\rm MoE}(x) & = & \sum_{i=1}^N g_i(x) \softmax(f_i(x)) \\
\label{eqn:moe-prob}
& = &\sum_{i=1}^N p(e_i|x) p(c|e_i,x) ~=~ p(c|x) 
\end{eqnarray}
This can also be seen as a
probability model, where the final probability over classes is marginalized
over the selection of expert: setting $p(e_i|x) = g_i(x)$ and $p(c|e_i,x) =
\softmax(f_i(x))$, we have \eqn{moe-prob}.

A product of experts (PoE) \cite{Hinton99} is similar, but instead combines log probabilities to form a product:
\begin{eqnarray}
F_{\rm PoE}(x) & \propto & \prod_{i=1}^N \softmax(f_i(x)) = \prod_{i=1}^N p_i(c|x)
\end{eqnarray}

Also closely related to our work is the Hierarchical Mixture of Experts
\cite{Jordan94}, which learns a hierarchy of gating networks in a tree
structure.  Each expert network's output corresponds to a leaf in the tree; the
outputs are then mixed according to the gating weights at each node.  

Our model differs from each of these three models because it dynamically assembles a suitable
expert combination for each input.  This is an instance
of the concept of \emph{conditional computation} put forward by Bengio \cite{Bengio13a} and examined in a single-layer stochastic setting by Bengio, Leonard and Courville \cite{Bengio13b}.  By conditioning our gating and expert
networks on the output of the previous layer, 
our model can express an exponentially large number of effective experts.

\section{Approach}
\vspace{-0.5mm}

\begin{figure}[t]
\vspace{-0.25in}
\centering
\begin{tabular}{M{2.5in}M{2.5in}}
\includegraphics[width=3in]{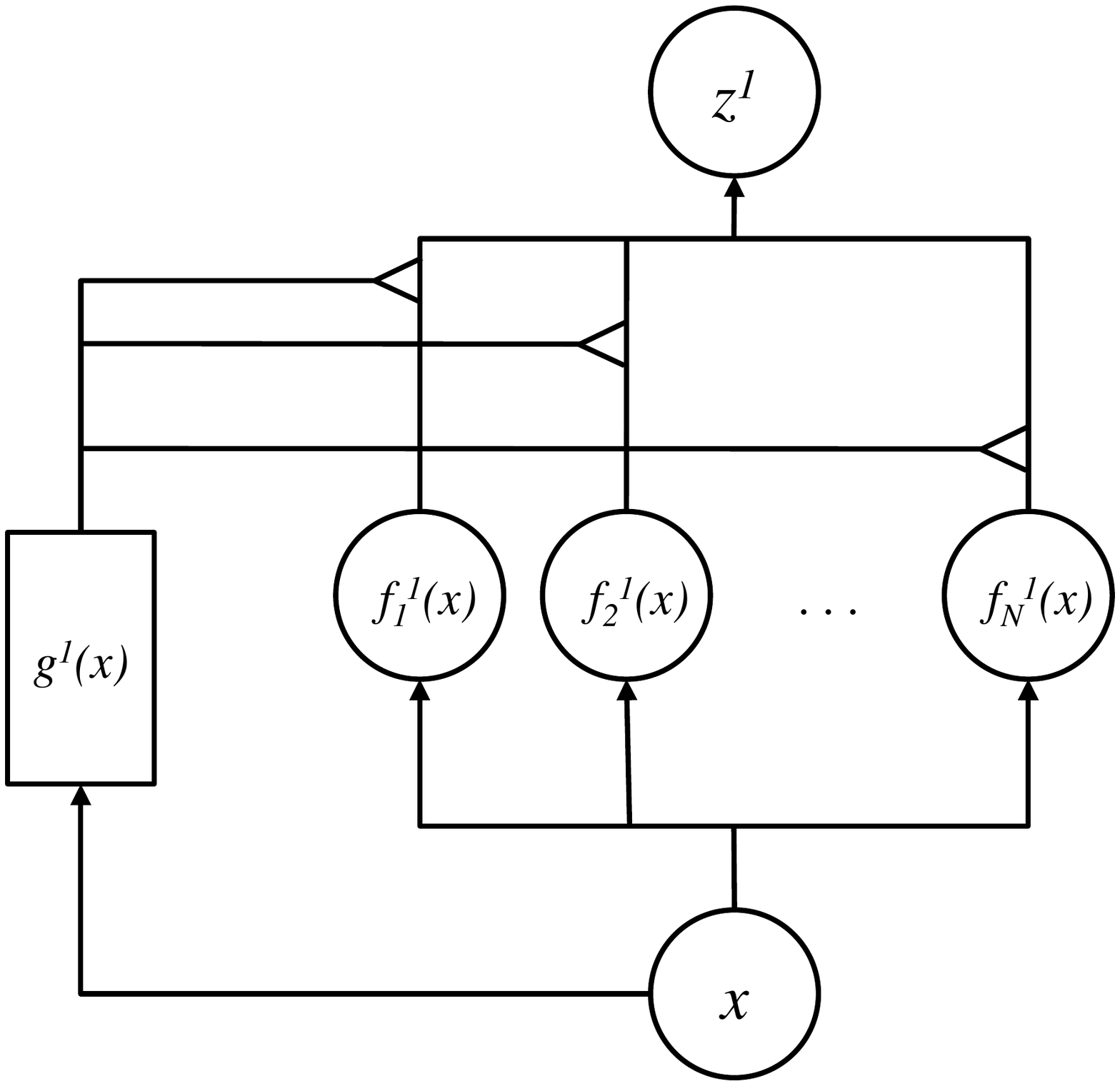}
&
\includegraphics[width=3in]{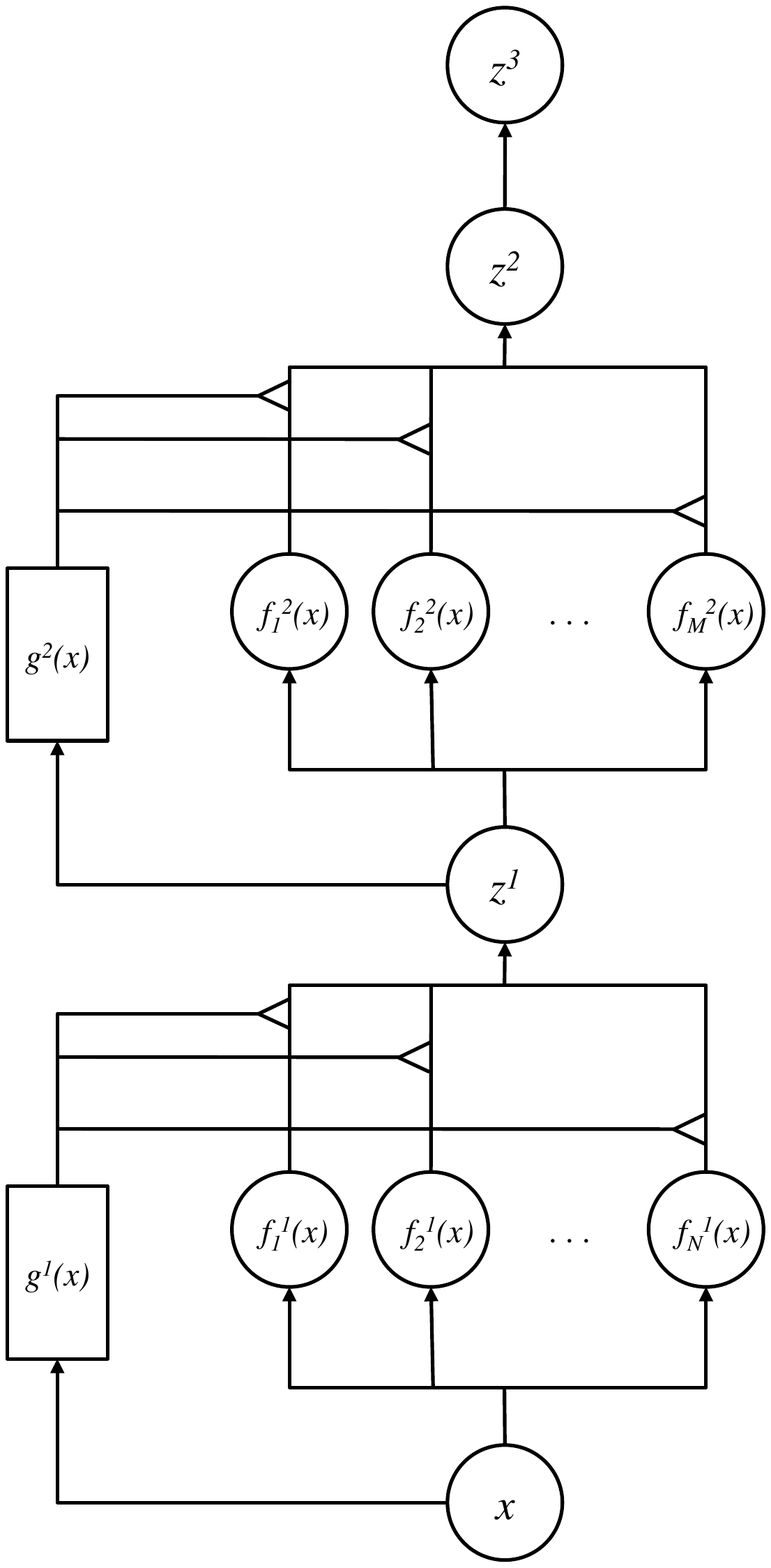}
\\
\vspace{-0.7in}
(a) &
\vspace{-0.7in}
(b) \\
\end{tabular}
\vspace{-8mm}
\caption{
(a) Mixture of Experts;
(b)
Deep Mixture of Experts with two layers.
}
\label{fig:arch-2layer}
\vspace{0mm}
\end{figure}

To extend MoE to a DMoE, we introduce two sets of experts with
gating networks $(g^1, f^1_i)$ and $(g^2, f^2_j)$, along with a final linear
layer $f^3$ (see \fig{arch-2layer}).  The final output is produced by composing
the mixtures at each layer:
\begin{eqnarray*}
z^1 & = & \sum_{i=1}^N g^1_i(x) f^1_i(x) \\
z^2 & = & \sum_{j=1}^M g^2_j(z^1) f^2_j(z^1) \\
F(x) ~ = ~~ z^3 & = & {\rm softmax}(f^3(z^2)) 
\end{eqnarray*}
We set each $f_i^l$ to a single linear map with rectification, and each $g_i^l$
to two layers of linear maps with rectification (but with few hidden units);
$f^3$ is a single linear layer.  See \secc{experiments} for details.

We train the network using stochastic gradient descent (SGD) with an additional
constraint on gating assignments (described below).  SGD by itself results in a
degenerate local minimum:  The experts at each layer that perform best for the
first few examples end up overpowering the remaining experts.  This happens
because the first examples increase the gating weights of these experts, which
in turn causes them to be selected with high gating weights more frequently.
This causes them to train more, and their gating weights to increase again,
{\it ad infinitum}.

To combat this, we place a constraint on the relative gating assignments to
each expert during training.  Let $G^l_i(t) = \sum_{t'=1}^t g^l_i(x_{t'})$ be
the running total assignment to expert $i$ of layer $l$ at step $t$, and let
$\bar{G^l}(t) = \frac1N \sum_{i=1}^N G^l_i(t)$ be their mean (here, $x_{t'}$ is
the training example at step $t'$).  Then for each expert $i$, we set
$g^l_i(x_t) = 0$ if $G^l_i(t) - \bar{G^l}(t) > m$ for a margin threshold $m$,
and renormalize the distribution $g^l(x_t)$ to sum to 1 over experts $i$.  This
prevents experts from being overused initially, resulting in balanced
assignments.  After training with the constraint in place, we lift it and
further train in a second fine-tuning phase.

\section{Experiments}
\label{sec:experiments}

\subsection{Jittered MNIST}

We trained and tested our model on MNIST with random uniform translations of
$\pm 4$ pixels, resulting in grayscale images of size $36 \times 36$.  As
explained above, the model was trained to classify digits into ten classes.

For this task, we set all $f^1_i$ and $f^2_j$ to one-layer linear models with
rectification, $f^1_i(x) = \max(0, W^1_ix + b^1_i)$, and similarly for $f^2_j$.
We set $f^3$ to a linear layer, $f^3(z^2) = W^3z^2 + b^3$.  We varied the
number of output hidden units of $f^1_i$ and $f^2_j$ between 20 and 100.  The
final output from $f^3$ has 10 units (one for each class).

The gating networks $g^1$ and $g^2$ are each composed of two linear+rectification layers with either 50
or 20 hidden units, and 4 output units (one for each expert), \ie $g^1(x) =
\softmax(B \cdot \max(0, A x + a) + b)$, and
similarly for $g^2$.

We evaluate the effect of using a mixture at the second layer by comparing
against using only a single fixed expert at the second layer, or concatenating
the output of all experts.  Note that for a mixture with $h$ hidden units, the
corresponding concatenated model has $N\cdot h$ hidden units.  Thus we expect
the concatenated model to perform better than the mixture, and the mixture to
perform better than the single network.  It is best for the mixture to be as
close as possible to the concatenated-experts bound.  In each case, we keep the
first layer architecture the same (a mixture).

We also compare the two-layer model against a one-layer model in which the
hidden layer $z^1$ is mapped to the final output through linear layer and
softmax.  
Finally, we compare against a fully-connected deep network with the
same total number of parameters.  This was
constructed using the same number of second-layer units $z^2$, but expanding
the number first layer units $z^1$
such that the total number of parameters is the same as the DMoE (including
its gating network parameters).

\subsection{Monophone Speech}

In addition, we ran our model on a dataset of monophone speech samples.  This
dataset is a random subset of approximately one million samples from a larger
proprietary database of several hundred hours of US English data collected
using Voice Search, Voice Typing and read data \cite{Jaitly12}.  For our
experiments, each sample was limited to 11 frames spaced 10ms apart, and had 40
frequency bins.  Each input was fed to the network as a 440-dimensional vector.
There were 40 possible output phoneme classes.

We trained a model with 4 experts at the first layer and 16 at the second
layer.  Both layers had 128 hidden units.  The gating networks were each two
layers, with 64 units in the hidden layer.
As before, we evaluate the effect of using a mixture at the second layer by
comparing against using only a single expert at the second layer, or
concatenating the output of all experts.

\section{Results}

\subsection{Jittered MNIST} \tab{error-mnist} shows the error on the training
and test sets for each model size (the test set is the MNIST test set with a
single random translation per image).  In most cases, the deeply stacked
experts performs between the single and concatenated experts baselines on the
training set, as expected.  However, the deep models often suffer from
overfitting: the mixture's error on the test set is worse than that of the
single expert for two of the four model sizes.  Encouragingly, the DMoE performs
almost as well as a fully-connected network (DNN) with the same number of
parameters, even though this network imposes fewer constraints on its structure.

\begin{figure}[h]
\vspace{-3mm}
\centering
{\bf Jittered MNIST:  Two-Layer Deep Model} \\
\begin{tabular}{|M{0.5in}|M{1.5in}M{2in}|}
\hline
& by Translation & by Class \\
\hline
Layer 1 & \includegraphics[width=1.5in]{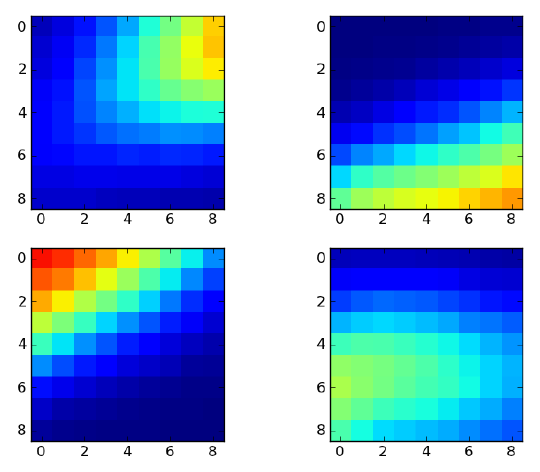} &
\includegraphics[width=2.0in]{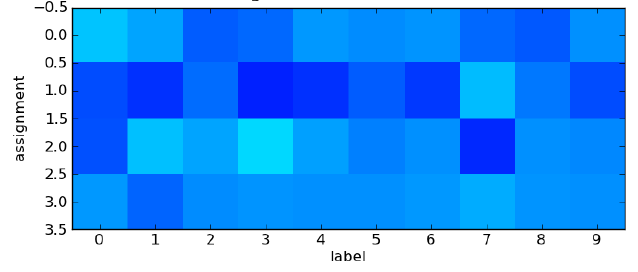} \\

\hline

Layer 2 & \includegraphics[width=1.5in]{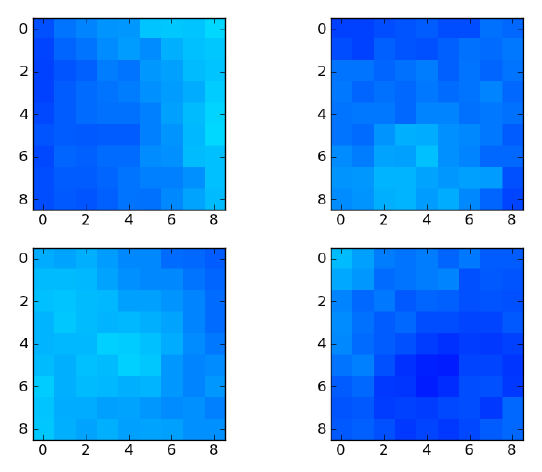} &
\includegraphics[width=2.0in]{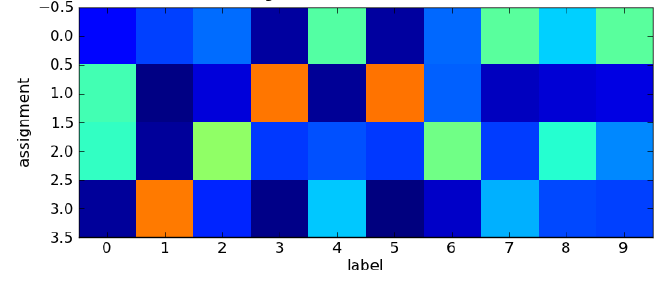} \\
\hline
\hline
1-Layer MoE without jitters & --- & \includegraphics[width=1.75in]{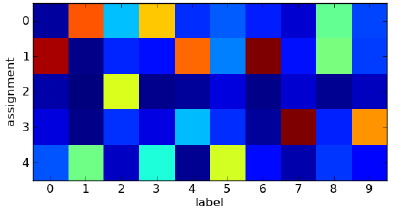} \\
\hline
\end{tabular}

\caption{
Mean gating output for the first and second layers, both by translation and by
class.  Color indicates gating weight.  The distributions by translation show
the mean gating assignment to each of the four experts for each of the
$9\times9$ possible translations.  The distributions by class show the mean
gating assignment to each of the four experts (rows) for each of the ten
classes (columns).  Note the first layer produces assignments exclusively by
translation, while the second assigns experts by class.  For comparison, we
show assignments by class of a standard MoE trained on MNIST without jitters,
using 5 experts $\times$ 20 hidden units.
}
\label{fig:jmnist_asgn}
\vspace{-3mm}
\end{figure}

In \fig{jmnist_asgn}, we show the mean assignment to each expert (\ie the mean
gating output), both by input translation and by class.  The first layer
assigns experts according to translation, while assignment is uniform by class.
Conversely, the second layer assigns experts by class, but is uniform according
to translation.  This shows that the two layers of experts are indeed being
used in complementary ways, so that all combinations of experts are effective.
The first layer experts become selective to \emph{where} the digit appears,
regardless of its membership class, while the second layer experts are
selective to \emph{what} the digit class is, irrespective of the digit's
location.

\begin{figure}[t]
\centering
\includegraphics[width=5.5in]{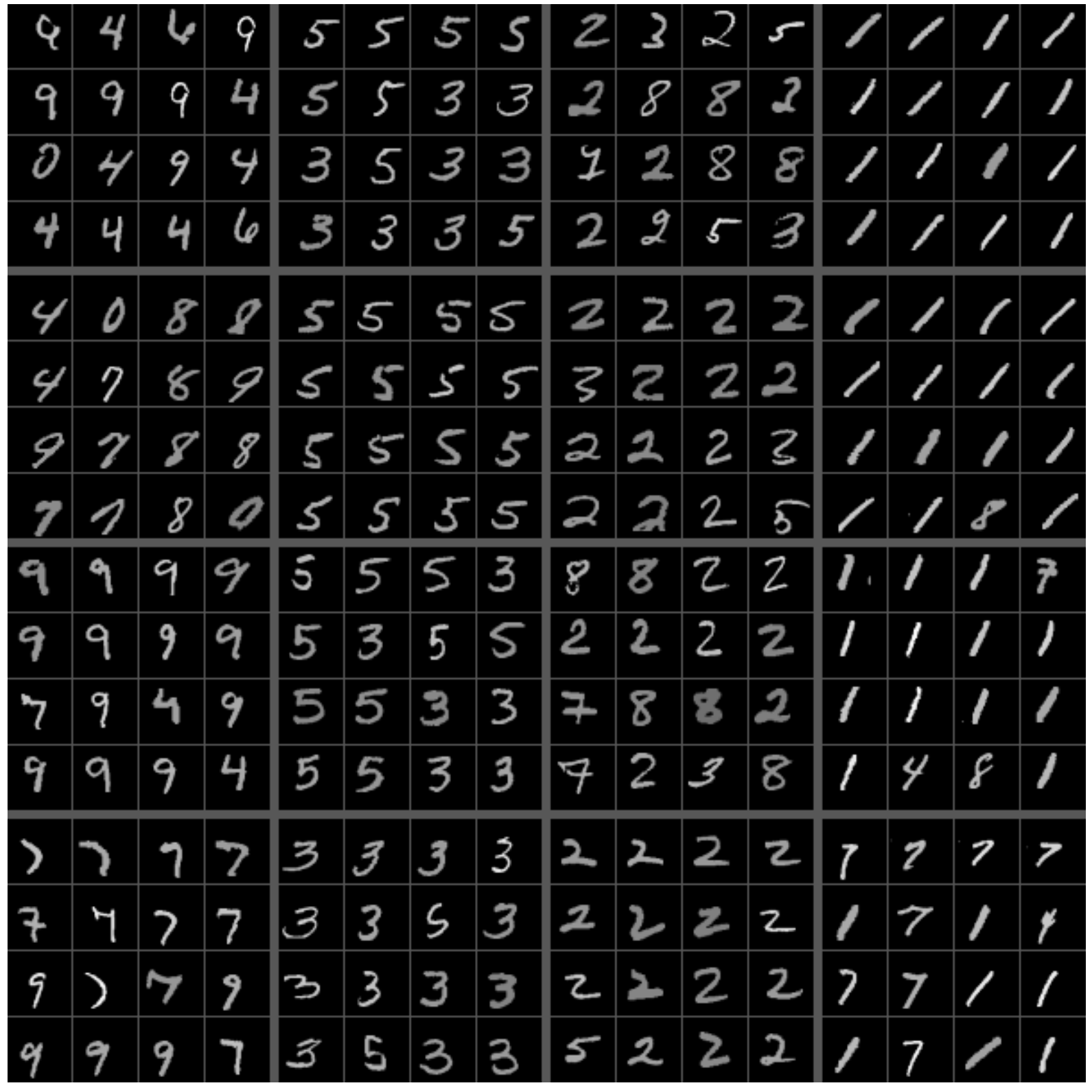}
\caption{\small
The nine test examples with highest gating value for each combination of experts, for the jittered mnist dataset.  First-layer experts are in rows, while second-layer are in columns.}
\label{fig:jmnist_samples}
\end{figure}

Finally, \fig{jmnist_samples} shows the nine test examples with highest gating
value for each expert combination.  First-layer assignments run over the rows,
while the second-layer runs over columns.  Note the translation of each
digit varies by rows but is constant over columns, while the opposite is true
for the class of the digit.  Furthermore, easily confused classes tend to be
grouped together, \eg 3 and 5.

\begin{table}[h]
\centering
{\bf Test Set Error:  Jittered MNIST} \\
\begin{tabular}{|l|l||c|c|c|c|}
\hline 
\bf Model & \bf Gate Hids & \bf Single Expert & \bf DMoE & \bf Concat Layer2 & \bf DNN \\
\hline \hline
$4 \times 100 - 4 \times 100$ & $50-50$ & 1.33 & 1.42 & 1.30 & 1.30 \\ 
\hline
$4 \times 100 - 4 \times 20$ & $50-50$  & 1.58 & 1.50 & 1.30 & 1.41 \\ 
\hline
$4 \times 100 - 4 \times 20$ & $50-20$  & 1.41 & 1.39 & 1.30 & 1.40 \\ 
\hline
$4 \times 50 - 4 \times 20$ & $20-20$   & 1.63 & 1.77 & 1.50 & 1.67 \\ 
\hline \hline
$4 \times 100$ (one layer) & 50         & 2.86 & 1.72 & 1.69 & -- \\
\hline
\end{tabular}

\vspace{0.15in}

{\bf Training Set Error:  Jittered MNIST} \\
\begin{tabular}{|l|l||c|c|c|c|}
\hline 
\bf Model & \bf Gate Hids & \bf Single Expert & \bf DMoE & \bf Concat Layer2 & \bf DNN \\
\hline \hline
$4 \times 100 - 4 \times 100$ & $50-50$  & 0.85 & 0.91 & 0.77 & 0.60 \\
\hline
$4 \times 100 - 4 \times 20$ & $50-50$   & 1.05 & 0.96 & 0.85 & 0.90 \\ 
\hline
$4 \times 100 - 4 \times 20$ & $50-20$   & 1.04 & 0.98 & 0.87 & 0.87 \\ 
\hline
$4 \times 50 - 4 \times 20$ & $20-20$    & 1.60 & 1.41 & 1.33 & 1.32 \\
\hline \hline
$4 \times 100$ (one layer) & 50          & 2.99 & 1.78 & 1.59 & -- \\
\hline
\end{tabular}
\caption{
Comparison of DMoE for MNIST with random translations, against baselines (i) using only one second layer expert, (ii) concatenating all second layer experts, and (iii) a DNN with same total number of parameters.  For both (i) and (ii), experts in the first layer are mixed to form $z^1$.  Models are annotated with ``\# experts $\times$ \# hidden units'' for each layer.
}
\label{tab:error-mnist}
\vspace{-2mm}
\end{table}

\subsection{Monophone Speech}
\vspace{-1mm}

\tab{error-monophone} shows the error on the training and test sets.  As was
the case for MNIST, the mixture's error on the training set falls between the
two baselines.  In this case, however, test set performance is about the same
for both baselines as well as the mixture.

\begin{figure}[h]
\centering
\includegraphics[width=5.5in]{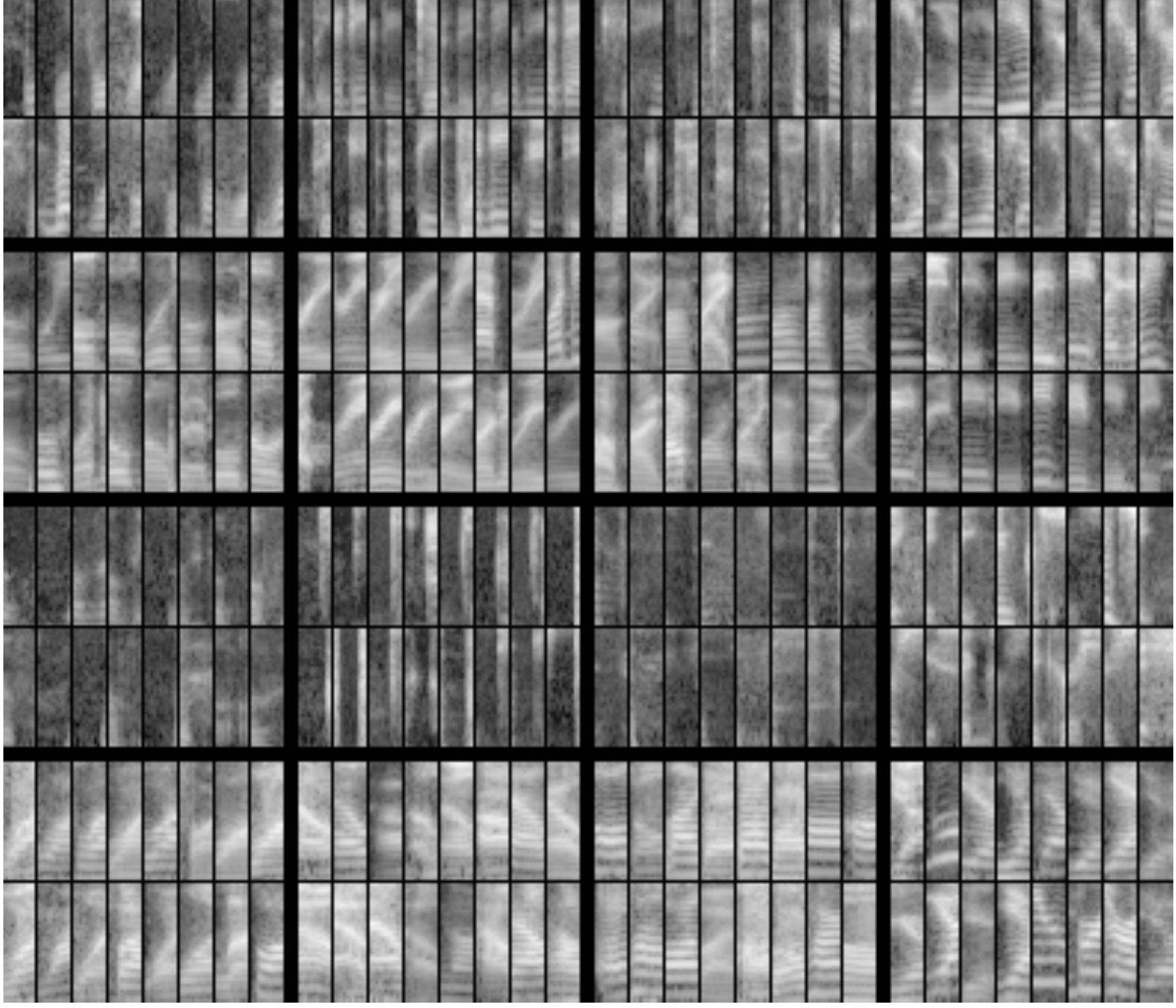}
\caption{\small
The 16 test examples with highest gating value for each combination of experts for the monophone speech data.  First-layer experts are in rows, while second-layer are in columns.  Each sample is represented by its 40 frequency values (vertical axis) and 11 consecutive frames (horizontal axis).  For this figure, we use four experts in each layer.}
\label{fig:monophone_samples}
\end{figure}

\begin{figure}[h]
\centering
{\bf Monophone Speech:  Conditional Assignments} \\
\includegraphics[width=3.0in]{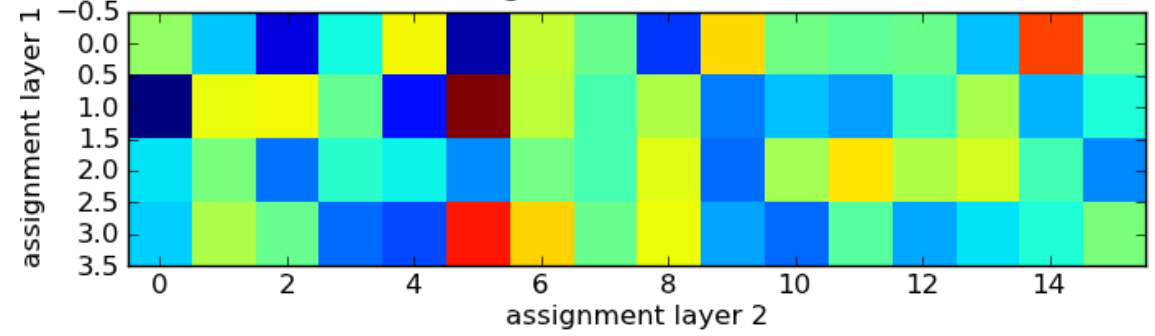}

\caption{
Joint assignment counts for the monophone speech dataset.  Here we plot the
average product of first and second layer gating weights for each expert combination.  We normalize each row, to produce a conditional distribution:  This shows the average gating assignments in the second layer given a first layer assignment.
Note the joint assignments are well mixed:  Choice of second layer expert is
not very dependent on the choice of first layer expert.  Colors range from
dark blue (0) to dark red (0.125).
}
\label{fig:monophone_asgn}
\end{figure}

\fig{monophone_samples} shows the 16 test examples with highest gating value
for each expert combination (we show only 4 experts at the second layer due to
space considerations).  As before, first-layer assignments run over the rows,
while the second-layer runs over columns.  While not as interpretable as for
MNIST, each expert combination appears to handle a distinct portion of the
input.  This is further bolstered by \fig{monophone_asgn}, where we plot the
average number of assignments to each expert combination.  Here, the choice
of second-layer expert depends little on the choice of first-layer expert.

\begin{table}[h]
\vspace{-1mm}
\centering
{\bf Test Set Phone Error:  Monophone Speech} \\
\begin{tabular}{|l|l||c|c|c|}
\hline 
\bf Model & \bf Gate Hids & \bf Single Expert & \bf Mixed Experts & \bf Concat Layer2 \\
\hline \hline
$4 \times 128 - 16 \times 128$ & $64-64$ & 0.55 & 0.55 & 0.56 \\
\hline \hline
$4 \times 128$ (one layer)  & $64$       & 0.58 & 0.55 & 0.55 \\
\hline
\end{tabular}

\vspace{0.15in}

{\bf Training Set Phone Error:  Monophone Speech} \\
\begin{tabular}{|l|l||c|c|c|}
\hline 
\bf Model & \bf Gate Hids & \bf Single Expert & \bf Mixed Experts & \bf Concat Layer2 \\
\hline \hline
$4 \times 128 - 16 \times 128$ & $64-64$  & 0.47 & 0.42 & 0.40 \\
\hline \hline
$4 \times 128$ (one layer)  & $64$        & 0.56 & 0.50 & 0.50 \\
\hline
\end{tabular}
\caption{
Comparison of DMoE for monophone speech data.  Here as well, we compare against baselines using only one second layer expert, or concatenating all second layer experts.
}
\label{tab:error-monophone}
\vspace{-5mm}
\end{table}

\clearpage

\section{Conclusion}

The Deep Mixture of Experts model we examine is a promising step towards
developing large, sparse models that compute only a subset of themselves for
any given input.  We see precisely the gating assignments required to make
effective use of all expert combinations: for jittered MNIST, a factorization
into translation and class, and distinctive use of each combination for
monophone speech data.  However, we still use a continuous mixture of the
experts' outputs rather than restricting to the top few --- such an extension is
necessary to fulfill our goal of using only a small part of the model for each
input. 
A method that accomplishes this for a single layer has been described
by Collobert \etal \cite{Collobert03}, which could possibly be adapted to our multilayer case; we hope to address this in future work.

\section*{Acknowledgements}

The authors would like to thank Matthiew Zeiler for his contributions on
enforcing balancing constraints during training.

\newpage

\vspace{0.5in}

{
\clearpage
\baselineskip=2pt
\bibliographystyle{plain}
\bibliography{smix}
}

\end{document}